\documentclass[11pt]{article}

\usepackage{acl}

\usepackage{times}
\usepackage{latexsym}
\usepackage{booktabs} 
\usepackage{longtable} 
\usepackage{multirow}
\usepackage[utf8]{inputenc}

\usepackage[T1]{fontenc}

\usepackage[utf8]{inputenc}
\usepackage{graphicx}

\usepackage{microtype}
\usepackage{amsmath}
\usepackage{amssymb}

\usepackage{inconsolata}

\title{Rectifying Demonstration Shortcut in In-Context Learning}

\author{
Joonwon Jang\(^{1}\) \ 
Sanghwan Jang\(^{2}\) \
Wonbin Kweon\(^{3}\) \
Minjin Jeon\(^{1}\)  \ 
Hwanjo Yu\(^{1,2,*}\)\\[1ex] 
Graduate School of AI, POSTECH\(^{1}\) \\
Department of Computer Science and Engineering, POSTECH\(^{2}\) \\
Institute of Artificial Intelligence, POSTECH\(^{3}\) \\
\texttt{\{kaoara, s.jang, kwb4453, minjinj, hwanjoyu\}@postech.ac.kr}
}

\begin{document}
\maketitle
\begin{abstract}
Large language models (LLMs) are able to solve various tasks with only a few demonstrations utilizing their in-context learning (ICL) abilities.
However, LLMs often rely on their pre-trained semantic priors of demonstrations rather than on the input-label relationships to proceed with ICL prediction. 
In this work, we term this phenomenon as the `Demonstration Shortcut'.
While previous works have primarily focused on improving ICL prediction results for predefined tasks, we aim to rectify the Demonstration Shortcut, thereby enabling the LLM to effectively learn new input-label relationships from demonstrations.
To achieve this, we introduce In-Context Calibration, a demonstration-aware calibration method.
We evaluate the effectiveness of the proposed method in two settings: (1) the Original ICL Task using the standard label space and (2) the Task Learning setting, where the label space is replaced with semantically unrelated tokens.
In both settings, In-Context Calibration demonstrates substantial improvements, with results generalized across three LLM families (OPT, GPT, and Llama2) under various configurations.
\begingroup
\renewcommand\thefootnote{}
\footnote{* Corresponding author}
\addtocounter{footnote}{-1}
\endgroup
\footnote{\url{https://github.com/Lainshower/In-Context-Calibration.git}}
\end{abstract}

\section{Introduction}


Large language models (LLMs) have demonstrated their effectiveness on a wide range of tasks through in-context learning (ICL), where models learn to perform a task from demonstrations \citep{brown2020language}. 
Leveraging their pre-trained knowledge, LLMs can associate various words in the demonstration with specific semantics (e.g., associating `extremely painful' with `negative'), thereby performing new tasks using only a small set of input-label examples, without requiring parameter updates \citep{dong2022survey, wei2023larger}.

However, LLMs often rely on the semantics from their pre-trained knowledge of given demonstrations, resulting in insufficient task learning for the patterns of the provided input-label pairs.
\citep{reynolds2021prompt, min-etal-2022-rethinking, wei2023larger, pan-etal-2023-context}.
This issue intensifies as the model size decreases \citep{wei2023larger}.
\citet{kossen2023context} suggest that smaller LLMs show promise in learning new mappings from demonstrations in some tasks, yet they still struggle to override semantic priors acquired during pre-training. 
Therefore, it is necessary to develop a method that enables LLMs of various sizes to effectively mitigate semantic priors preferences and learn to perform unseen tasks from demonstrations.

Prior works have primarily focused on the instabilities of LLMs in ICL prediction \citep{holtzman-etal-2021-surface, fei2023mitigating}.
To mitigate these instabilities, these studies introduced content-free tokens or utilized the entire test set to calibrate prediction probabilities \citep{holtzman-etal-2021-surface, fei2023mitigating, zhou2023batch}.
However, they lack consideration of the semantic priors of LLMs on the demonstration and do not verify whether their approach enhances LLMs to learn new tasks from the demonstrations.

\begin{figure*}[ht!]
  \centering
  \includegraphics[width=\textwidth]{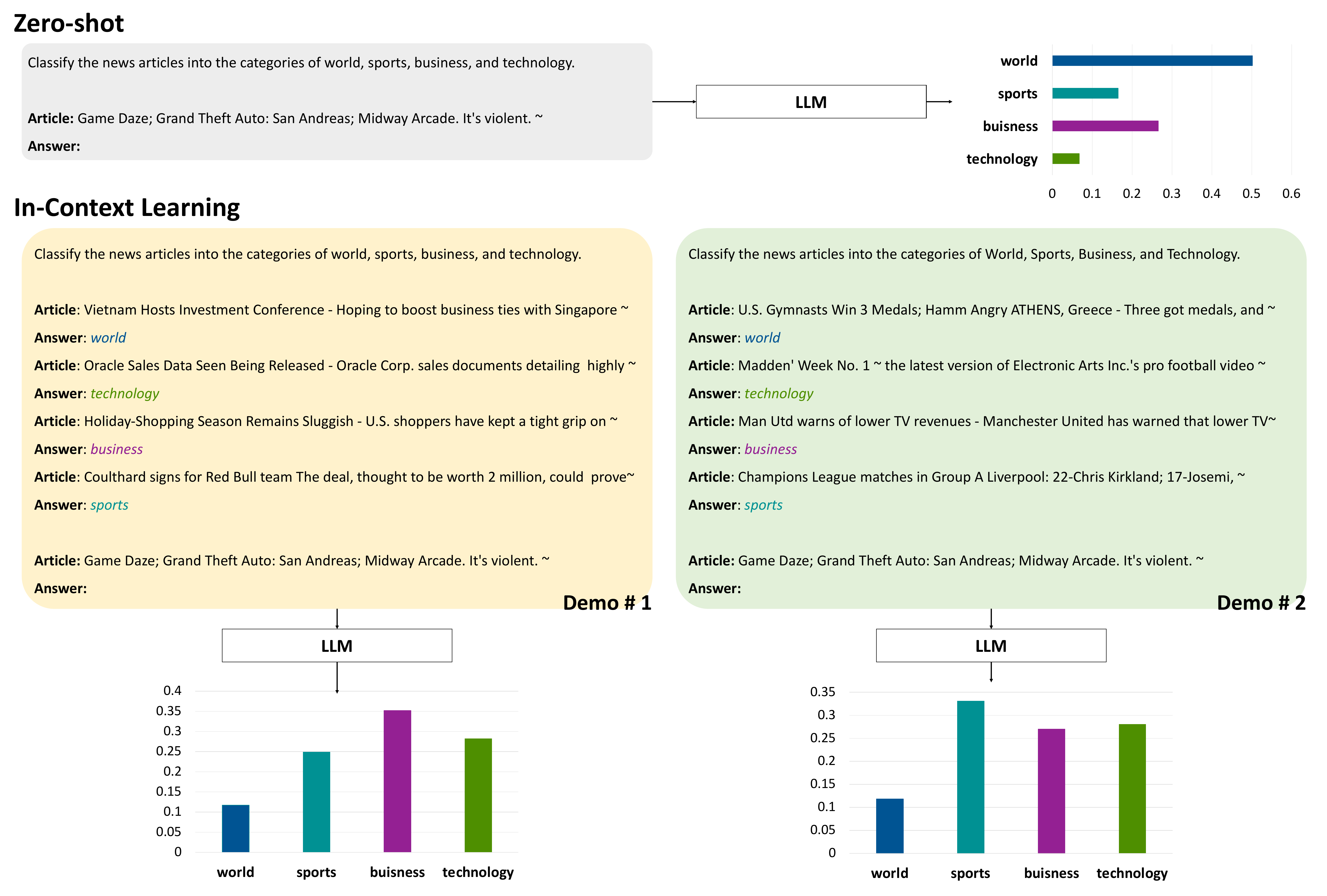}
    \caption{The overall illustration of the \textit{Demonstration Shortcut}. In a zero-shot setting, an LLM predicts the test article to the world label. With the first demonstration set, the LLM predicts the business label through ICL. However, with the second demonstration set — which has the same label order but different semantics in the examples — the LLM predicts the sports label. GPT-J is used for these experiments. See Appendix \ref{appendix:full_description} for a full description of the demonstrations.
  }
  \label{fig:short-cut}
\end{figure*}

In this work, we investigate how the LLMs' pre-trained knowledge on the demonstrations affects ICL.
We define the following phenomenon as a \textit{Demonstration Shortcut}: the reliance of LLMs on their pre-trained semantic priors of demonstrations in ICL prediction, rather than learning from the input-label relationships presented in these demonstrations.
Due to the Demonstration Shortcut, LLMs' ICL predictions may be overly dependent on the semantics of the given demonstrations even when the label distribution is uniform and the order is identical (Figure \ref{fig:short-cut}).

To tackle this problem, we propose In-Context Calibration, a method designed to rectify the Demonstration Shortcut in ICL. 
In-Context Calibration estimates the semantic prior of LLMs on each demonstration sample with the in-context examples. 
Formally, for each example in the demonstration, we estimate its semantic prior relative to the remaining examples and calculate the expected semantic priors of the demonstrations.
At test time, we use this term to rectify LLMs' dependency on semantic priors and enable the model to learn the intended input-label relationships from the demonstrations.

We evaluate the effectiveness of In-Context Calibration on 27 classification datasets from two perspectives: (1) Original ICL Task and (2) Task Learning settings. 
In the Original ICL Task, we use the standard label space.
In the Task Learning setting, the label space is replaced with semantically unrelated tokens.
This requires LLMs to learn the novel input-label relationships to achieve high performance, as these relationships are never seen in pre-training.
Our proposed method not only demonstrated enhanced performance across various tasks but also showed improvement in task learning abilities.
Specifically, In-Context Calibration outperforms other ICL methods in Natural Language Inference (NLI) tasks, which demand high task learning ability.
We also demonstrate that In-Context Calibration enhances ICL performance across various model types and sizes, effectively rectifying the `Demonstration Shortcut' problem.


\section{Backgrounds}




\paragraph{How LLMs utilize demonstrations in ICL} 
Following ICL's accomplishments, extensive prior works have sought to understand how LLMs use demonstrations, yet there is still no consensus on the following two contradictory perspectives.
One line of research claims that LLMs do not learn new input-label relationships from the demonstrations, with the evidence that ICL performance only marginally drops when labels in the demonstrations are replaced with random labels \citep{min-etal-2022-rethinking}.
Instead, LLMs independently recognize the semantics of input and label of in-context demonstrations using their pre-trained knowledge and perform ICL prediction with its language modeling objective \citep{reynolds2021prompt, min-etal-2022-rethinking}.
On the other hand, while some studies suggest that LLMs can learn novel tasks through demonstrations, there is a notable lack of concrete experimental proof in real-world LLM applications \citep{xie2021explanation, zhang2023and}.
Addressing this gap, \citet{wei2023larger} provides evidence that larger LLMs can learn the new input-label mappings from demonstrations.
Summarizing these perspectives, \citet{pan-etal-2023-context} show that applying pre-trained knowledge to demonstrations for task recognizing is a broad capability across scales while learning new input-label mappings becomes more feasible as the scale increases. 
This indicates that as LLMs decrease in size, they rely more on pre-trained knowledge of demonstrations in ICL prediction. 
Furthermore, when labels in the demonstrations are flipped with different labels (e.g., labeling `positive' as `negative' and vice versa), these models struggle to override semantic priors obtained during pre-training. \citep{wei2023larger, kossen2023context}.

\paragraph{Improving ICL through Calibration} 
Various studies have focused on the instability of ICL prediction in LLMs.
\citep{pmlr-v139-zhao21c, jiang2023generative} reveal the instability of ICL prediction arises from the majority label bias and recency label bias, and \citet{fei2023mitigating} identifies domain as a factor contributing to label bias behind this instability. 
These studies have attempted to estimate the instability of ICL prediction by introducing content-free tokens \citep{pmlr-v139-zhao21c} or using the entire test set to calibrate ICL prediction probabilities \citep{fei2023mitigating, zhou2023batch}.
Although their approaches show an improvement in ICL prediction, they do not address the reliance of LLMs on the semantic priors of the demonstration.
In other words, the primary objective of these methods is to enhance ICL prediction performance for pre-defined tasks, rather than enabling the model to learn input-label mappings from the demonstrations.
Furthermore, they fail to demonstrate whether their calibration method allows LLMs to learn new input-label mappings through demonstrations.
Moreover, it is unreasonable to assume that the entire test set is available when learning new input-label relationships.
Based on the discussions above, our analysis focuses on the reliance of LLMs on their semantic priors of demonstrations, offering a novel perspective on the calibration method.

\section{Demonstration Shortcut}

In this section, we first introduce a new typology termed \textit{Demonstration Shortcut}. 
This concept refers to the reliance of LLMs on their pre-trained semantic priors of demonstrations for making ICL predictions, rather than learning the input-label relationships presented in the demonstrations.
Figure \ref{fig:short-cut} illustrates the underlying concept of the Demonstration Shortcut.
In a zero-shot setting, the LLM predicts the test article to the world label (the ground-truth label is technology).
Next, we constructed two demonstration sets from the same training dataset, each with a uniform label distribution.
While these sets have identical label orders, they differ in the semantics of the examples.
The first demonstration set mostly features business-related semantics across all examples, while the second set leans more toward sports-related semantics.
After ICL with the first demonstration set, the LLM predicted the article to the business label.
However, LLM predicted the article to the sports label with the second demonstration set.
Despite both sets having uniform label distributions and identical orders, LLM relies on the semantic prior from each demonstration set, exhibiting a Demonstration Shortcut and failing to predict the correct answer.
This indicates that an over-dependence on semantic prior interrupts the ability of LLMs to learn the new input-label mapping relationships from the demonstrations.

\begin{table}[t!]
\resizebox{\columnwidth}{!}{
\centering 
\begin{tabular}{@{}lcccc@{}}
\toprule
           & Demo \#1 & Demo \#2 & Demo \#3 & Demo \#4 \\ \midrule
World      & 0.20   & 0.22   & \textbf{0.26}   & 0.17   \\
Sports     & 0.25   & \textbf{0.30}   & 0.25   & 0.25   \\
Business   & \textbf{0.34}   & 0.28   & 0.25   & 0.27   \\
Technology & 0.19   & 0.18   & 0.21   &\textbf{0.28}   \\ \bottomrule
\end{tabular}
}
\caption{Prediction distributions of GPT-J with different demonstration semantics sets. All four demonstration sets have uniform and identical label distribution.}
\label{table:shortcut}
\end{table}

To deeply describe how semantic priors acquired from pre-training affect ICL prediction, consider the first example in Demo \#1 of Figure \ref{fig:short-cut}, titled \textit{`Vietnam Hosts Investment Conference - Hoping to Boost Business Ties with Singapore $\sim$'} (see Appendix \ref{appendix:full_description} for the full example). 
The overall context of the article allows the LLM's pre-training knowledge to associate its semantics with the business label. 
Meanwhile, word-by-word examination (e.g., Vietnam or Singapore) also reveals potential associations of its pre-trained semantics with the world label \citep{Tang2023LargeLM}.
By iterating this process for all examples, regardless of the assigned ground-truth label, the LLM may proceed with ICL prediction based on pre-trained semantic distributions of the demonstrations, leading to the Demonstration Shortcut.

Substantiating our intuition, we conducted additional experiments based on 
Figure \ref{fig:short-cut}.
We constructed Demo \#3 to be characterized by world semantics across all examples, while Demo \#4 features across technology-related semantics.
All demonstration sets were designed to have a uniform label distribution and an identical sequence, as depicted in Figure \ref{fig:short-cut}.
We ensured the test set also followed a uniform label distribution (25 samples for each label) and reported the average label prediction probabilities for these examples in Table \ref{table:shortcut}.
Despite having uniform and identical label distributions in the demonstrations, the LLM predictions for all demonstration sets exhibit different behaviors; this variance aligns with the overall semantics of each demonstration.


\section{In-Context Calibration}

This section introduces a novel calibration method to rectify the Demonstration Shortcut. 
We first revisit existing calibration methods designed to improve ICL predictions and analyze their limitations, particularly regarding Demonstration Shortcut. 
We then propose In-Context Calibration, our approach to rectifying the Demonstration Shortcut in ICL.

\paragraph{Revisiting Previous Methods} Prior works on calibrating LLMs attempt to adjust ICL predictions by estimating the prompt prior with a content-free token `N/A' \citep{pmlr-v139-zhao21c} or by estimating the task prior by utilizing the entire test distribution \citep{fei2023mitigating, zhou2023batch}.
Specifically, Contextual Calibration (CC) estimates the content-free prediction prior as $P_{LM}(y|`N/A\textrm', [(x_i, y_i)]_{i \in [K]})$, where $x_i$ represents the text input, $y_i$ corresponds to a verbalized label name, and $K$ denotes the total number of examples.
Meanwhile, Domain Calibration (DC) estimates task prior with $\frac{1}{M}\sum_{r=1}^{M}P_{LM}(y|[r/w]_r, [(x_i, y_i)]_{i \in [K]})$, where $[r/w]$ represents random words drawn from the entire test set.
However, the introduced terms are not entirely content-free; their neutrality depends on the task type and the demonstrations \citep{fei2023mitigating, zhou2023batch}.
Therefore, if these terms' pre-trained semantics mismatch with the semantic priors of demonstrations, they are limited in rectifying the Demonstration Shortcut. 
Moreover, relying on the entire test distribution is impractical in real-world settings.

\paragraph{In-Context Calibration} To overcome these limitations, we propose In-Context Calibration, which rectifies the Demonstration Shortcut of the model in the ICL setting.

For each $x_i$  in the demonstration, we estimate the semantic prior of LLMs on each demonstration sample with remaining in-context demonstrations:

{\small\begin{equation}
P_{i} = P_{LM}(y \mid x_i, [(x_j, y_j)]_{j \in [K]\setminus\{i\}}), \\ 
 \forall i,
\end{equation}}

\begin{figure*}[ht!]
  \centering
  \includegraphics[width=0.97\textwidth]{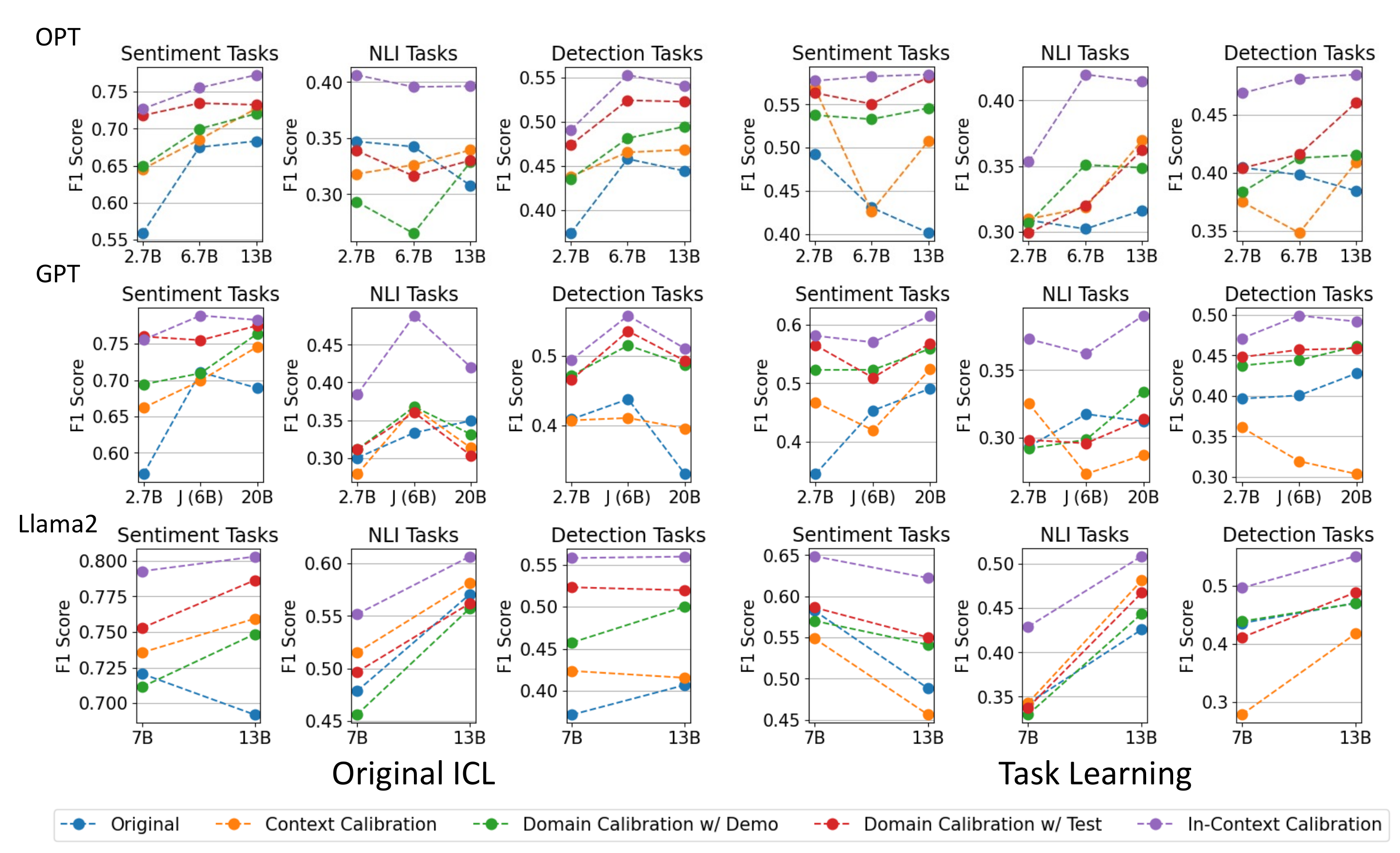}
  \caption{Averaged Macro F1 scores for OPT (Top), GPT (Medium), and Llama2 (Bottom) across Sentiment, NLI, and Detection Tasks. The left the left three columns depict the performance on the Original ICL Task. The right three columns plot the Task Learning scores. In both graphs, the x-axis represents the model size.
  }
  \label{fig:main_results}
\end{figure*}

\noindent where $P_i$ represents the semantic distribution of $x_i$ given the remaining $K-1$ demonstrations. 
This allows us to estimate the contextual semantic prior of each demonstration sample while preserving the remaining in-context demonstrations order and conditions present in the original ICL setting. 

Additionally, we estimate each demonstration sample's word-by-word semantic distribution by applying the random shuffling function $R()$ to $x_i$, which shuffles the order of the words as follows:


{\small\begin{equation}
P_{R(i)} = P_{LM}(y \mid R(x_i), [(x_j, y_j)]_{j \in [K]\setminus\{i\}}), \\ \forall i.
\end{equation}}

\noindent The resulting random order of $R(x_i)$ is not grammatically meaningful, yet it enables context-agnostic estimation of the LLMs' pre-trained semantics for each demonstration sample.
In other words, $R(x_i)$ is stripped of context and retains only the words with their meanings, thereby preventing the LLMs from making predictions based solely on the semantics of individual words \citep{fei2023mitigating, Tang2023LargeLM} \footnote{Please refer to Appendix \ref{appendix:random-shuffling} for additional details for the random shuffling function.}

We iterate this process across all $K$ demonstrations and compute the average to estimate the semantic priors of the demonstrations:

{\small\begin{equation}
\begin{aligned}
& \frac{1}{K}\sum_{i=1}^{K} (\lambda \cdot P_{i} + (1-\lambda) \cdot P_{R(i)}),
\end{aligned}
\end{equation}}

\noindent where $\lambda$ is the hyperparameter that controls the balance between the two terms.  Considering every dependency from the $K$ demonstrations and taking the average, we can estimate the expected semantic priors of demonstrations, enabling more effective \textit{demonstration-aware calibration}.
The model then makes ICL predictions based on the following estimates:

\begin{equation}
\small
\begin{split}
\hat{y}_{pred} &= \arg\max_{y \in \mathcal{L}} \frac{P_{LM}(y \mid x_{\text{test}}, [(x_i, y_i)]_{i \in [K]})}
{\frac{1}{K}\sum_{i=1}^{K} (\lambda \cdot P_{i} + (1-\lambda) \cdot P_{R(i)})},
\end{split}
\end{equation}

\noindent where ${P_{LM}(y \mid x_{\text{test}}, [(x_i, y_i)]_{i \in [K]})}$ is the original ICL prediction.

\section{Experimental Setups}

We investigate the effectiveness of In-Context Calibration in two aspects: (1) the model's performance on the Original ICL Task  (using standard label space from the dataset) and (2) the Task Learning setting (label space is randomly mapped to semantically unrelated tokens), following the experimental settings of \citet{pan-etal-2023-context}.
Since the newly introduced input-label mappings have never been parameterized during pre-training, the model must utilize its task learning abilities to handle the problem.
In our main experiment, unless stated otherwise, we conduct task learning experiments using string numbers\footnote{String numbers demonstrate better task learning ability than other tokens in \citet{pan-etal-2023-context}.}, and to avoid any pre-trained bias, each label is randomly assigned to a unique string number for every seed.

\subsection{Datasets}
We conducted experiments on 27 classification datasets across three types of tasks: Sentiment, NLI, and Detection classification task.
Our dataset selection and prompts largely follow the methodologies of prior ICL works \citep{pmlr-v139-zhao21c, min-etal-2022-rethinking, fei2023mitigating}, and more details are described in Appendix \ref{appendix:data_info}.

\subsection{Base Models}

To validate our method across a diverse set of models, we use three state-of-the-art LLM families: GPT (2.7B, J (6B), 20B) \citep{brown2020language}, OPT (2.7B, 6.7B, 13B) \citep{zhang2022opt}, and Llama2 (7B, 13B) \citep{touvron2023llama}.
For the GPT models, we use the open-sourced versions provided by EleutherAI \cite{gao2020pile, mesh-transformer-jax, black2022gpt} as \citet{fei2023mitigating}.
Consequently, we utilize checkpoints from the Transformers library \cite{wolf2020huggingfaces} for all the aforementioned models.

\subsection{Implementation Details}

Adopting a sampling-based evaluation approach, we sample different sets of demonstrations from the training set for each seed and report the mean and standard deviation of the results.
We use $K = 8$ examples and conduct five evaluations using different random seeds, per the methodology described by \citet{fei2023mitigating}.
Unless stated otherwise, we set $\lambda$ to $0.5$.
For the baselines, we selected Contextual Calibration (CC) \citep{pmlr-v139-zhao21c} and Domain Calibration (DC) \citep{fei2023mitigating} to assess the performance on the Original ICL Task and Task Learning setting.
For Domain Calibration, the original method involves constructing a bag of words from the entire test set, which is impractical for real-world inference.
To facilitate a fair comparison, we adapted this method.
The adapted version samples random words from the demonstration and is labeled as ``w/ Demo'' (with Demo), while we refer to the original method as ``w/ Test'' (with Test).

\section{Experimental Results}
\subsection{Main Results}
Figure \ref{fig:main_results} shows the Macro F1-scores of OPT, GPT, and Llama2 on three categories of datasets in the Original ICL Task and Task Learning setting.
The results demonstrate that In-Context Calibration effectively rectifies the Demonstration Shortcut, enhancing performance in the Original ICL Task and improving learning capabilities for new input-label tasks.
Specifically, for Llama2, In-Context Calibration resulted in an average F1 score improvement of 23\% compared to the original inference in the Original ICL Task.
Regarding CC using `N/A' token and DC w/ Test sampling random words from the test set, if the newly introduced tokens fail to accurately estimate the neutrality of the demonstration's semantic distribution, the model remains constrained by the Demonstration Shortcut, limiting performance improvements. 
This is particularly pronounced in NLI tasks, where previous works \citep{pan-etal-2023-context,kossen2023context} have shown to be challenging for LLMs to learn input-label pairs due to strong reliance on semantic priors in the ICL setting, while In-Context Calibration effectively rectifies the Demonstration Shortcut for all LLMs, resulting in notable performance increases.

This improvement is also evident in the Task Learning setting, where the label space is replaced with string numbers. 
For GPT, In-Context Calibration achieved an average F1 Score increase of 27\% over the original inference.
Particularly in the NLI task, other methodologies struggled to mitigate the dependency of the semantic priors of the models on demonstration, leading to decreased performance compared to original inference in some cases.
Across various dataset categories and model types, In-Context Calibration consistently outperformed baseline methods on tasks with novel input-label pairs by effectively reducing the model's reliance on the demonstration's semantic prior (see Appendix \ref{appendix:more_results} for comprehensive results).

\subsection{Ablation Study}

\begin{table}[h!]
\renewcommand{\arraystretch}{1.3} 
\resizebox{\columnwidth}{!}{
\centering
{\fontsize{9pt}{9pt}\selectfont 
\begin{tabular}{lcc}
\hline
Method & Original & TL \\
\hline
Original Inference & 0.48 & 0.40 \\
$ R(x_i) \xrightarrow{} `N/A\textrm' $  & 0.49 & 0.38 \\
$R(x_i) \xrightarrow{}$ random words & 0.57  & 0.46 \\
In Context Calibration ($\lambda = 0.5 $)& 0.60 & 0.49 \\
In Context Calibration ($\lambda = 0 $) & 0.56 & 0.46 \\
In Context Calibration ($\lambda = 1 $) & 0.58 & 0.50 \\
\hline
\end{tabular}
}}
\caption{\label{citation-guide}
Analysis of In-Context Calibration: Performance comparing $R(x_i)$ replacement with `N/A' (CC) or randomly sampled test set words (DC), and effects of varying $\lambda$ values, shown through average Macro F1 scores across 27 datasets in Original ICL Task (Original) and Task Learning (TL) settings using GPT-J.
}
\end{table}

We conducted an ablation study of the proposed method. 
First, we replaced the $R(x_i)$ term with either a `N/A' token or words randomly sampled from the bag of words constructed from the entire test set. 
The replacement with the `N/A' token resulted in lower performance compared to the original inference in the Task Learning setting, likely due to the instability of the `N/A'  token in estimating prompt neutrality \citep{fei2023mitigating}. 
Furthermore, words randomly sampled from the test set underperformed in both tasks compared to In-Context Calibration with $\lambda = 0.5$.
This underperformance could stem from the semantic distribution mismatch between the sampled words and the demonstrations, limiting the model's ability to rectify the Demonstration Shortcut. 
However, the demonstration-aware calibrating of In-Context Calibration leads to performance improvements.

\begin{figure*}[ht!]
  \centering
  \includegraphics[width=0.97\textwidth]{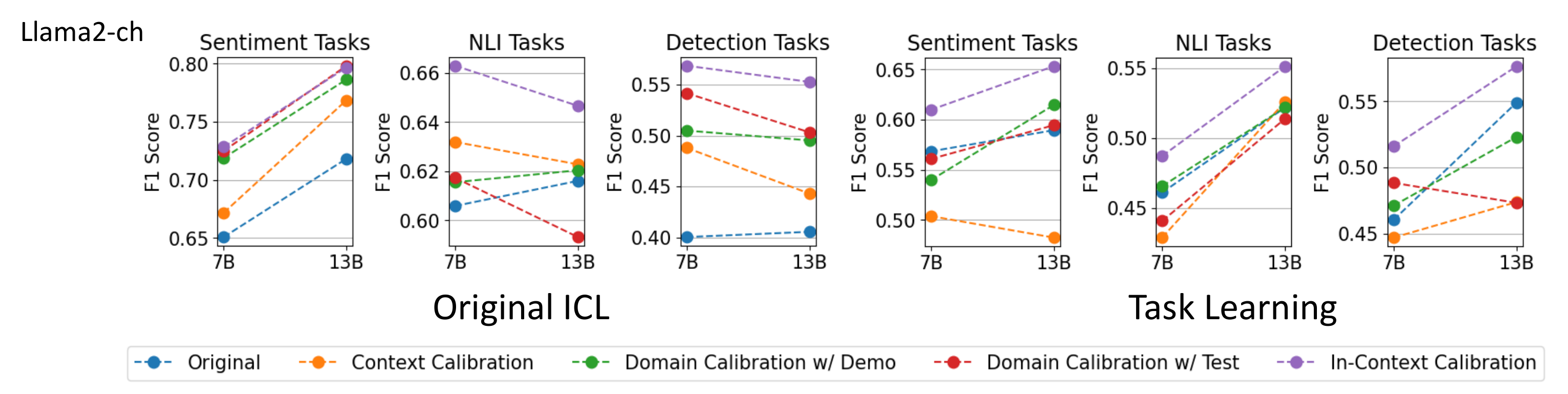}
  \caption{Averaged Macro F1 scores for Llama2-Chat across Sentiment, NLI, and Detection Tasks. The left three columns depict the performance of the Original ICL Task. The right three columns plot the Task Learning scores. In both graphs, the x-axis represents the model size.}
  \label{fig:llama2_chat}
\end{figure*}

Ablation studies on $\lambda$ demonstrate that using only $R(x_i)$ for calibration ($\lambda = 0$), which shuffles the token order for each demonstration, leads to limited performance gains in both the Original ICL Task and Task Learning setting, attributing to the loss of contextual information. 
On the other hand, not utilizing $R(x_i)$ for calibration ($\lambda = 1$) results in better learning of new input-label relationships. 
However, this approach is less effective in mitigating the model's dependency on word-wise pre-trained semantics of demonstration when considering labels in the Original ICL Task. 
We conduct a detailed task-wise analysis of the $\lambda$ value's effect in Section \ref{task-wise} and Appendix \ref{appendix:full-lambda}.
Therefore, after comprehensive analysis for $\lambda$ values, we set $\lambda$ to $0.5$ in the main experiment.\footnote{Please refer to the Appendix \ref{appendix:full-lambda} for the comprehensive analysis for $\lambda$ value.}

\subsection{Analysis with Enhanced Models}

\paragraph{Instruction-Tuned Model}
In previous studies \citep{min-etal-2022-rethinking, wei2023larger}, instruction tuning has been demonstrated to strengthen the usage of semantic priors in the demonstration. 
Therefore, we conducted experiments to determine whether In-Context Calibration consistently rectifies the Demonstration Shortcut and improves task learning ability in instruction-tuned LLMs. 
As depicted in Figure \ref{fig:llama2_chat}, In-Context Calibration increases the model's F1 score across all Original ICL Tasks and particularly shows substantial improvement over other calibration methods in the Task Learning setting. 
Especially in the Task Learning setting, other calibration methods often underperformed the original inference. 
These experiments demonstrate that In-Context Calibration remains effective and enhances the model's task learning abilities, even after instruction tuning has strengthened the LLM's reliance on semantic priors.

\paragraph{Over 50B Scale Models} 

\begin{figure*}[h!]
  \centering
  \includegraphics[width=0.7\textwidth]{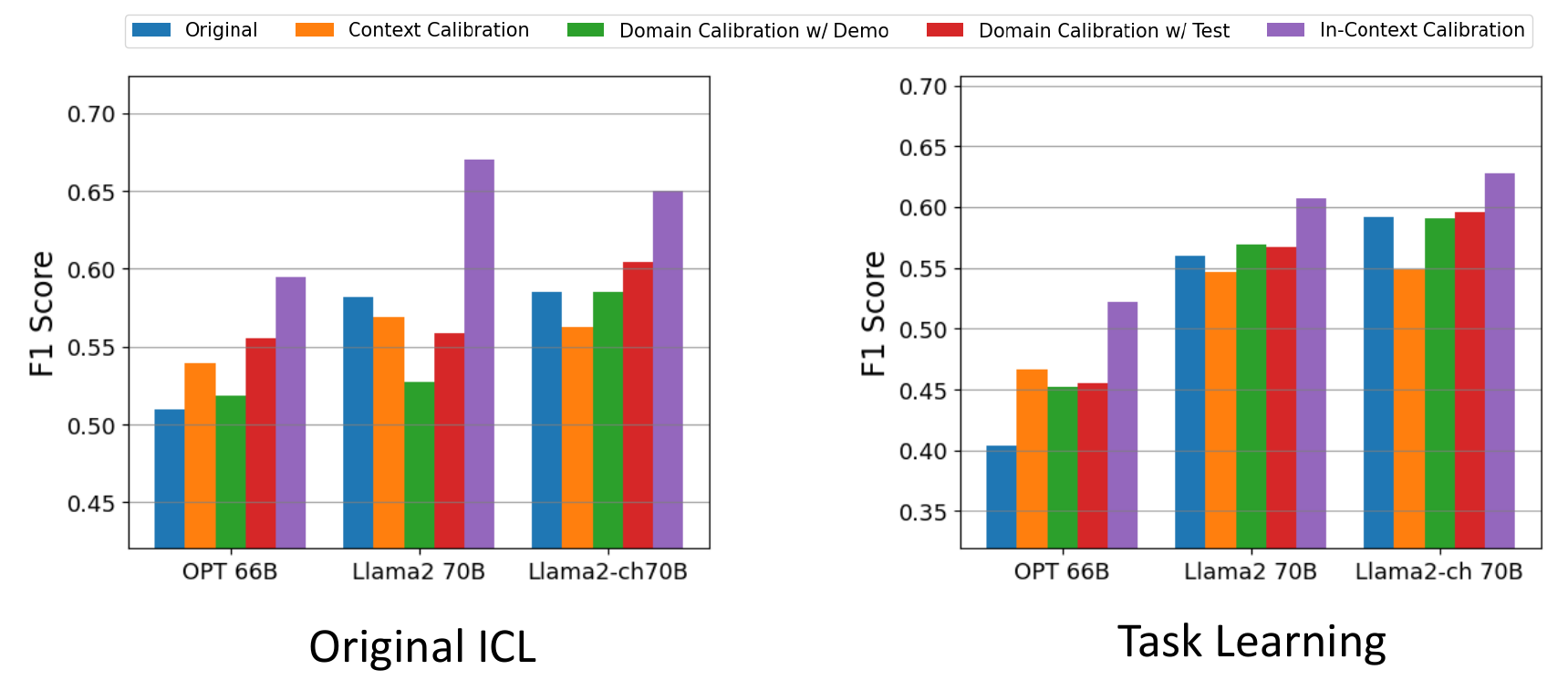}
  \caption{Averaged Macro F1 scores across 27 classification tasks for over 50B scale LLMs. The left graphs depict performance in the Original ICL Task, while the right graphs plot task learning scores. In both sets of graphs, the x-axis denotes the model type. Full details of the data-type scores are provided in Appendix \ref{appendix:more_results}.
  }
  \label{fig:large_model}
\end{figure*}

We conducted experiments to verify that the proposed method consistently improves performance in larger models. 
As demonstrated by \citet{pan-etal-2023-context}, in the Task Learning setting, the performance of some models (e.g., OPT) starts to match their performance in the Original ICL Task for models larger than 50B. 
These models can utilize the mapping information provided in the demonstrations. 
Therefore, we conducted experiments on OPT 66B, Llama2 70B, and Llama2-chat 70B, reporting average F1 scores for 27 datasets in both the Original ICL Task and Task Learning settings, as shown in Figure \ref{fig:large_model}. 
Particularly for Llama2 70B, In-Context Calibration improved F1-score performance by 15\% over the original inference in the Original ICL Task and by 9\% in the Task Learning setting, while some methods hurt the model’s performance.
These findings suggest that In-Context Calibration not only boosts performance in smaller-scale models but also consistently improves the learning ability of larger-scale models, by effectively rectifying the Demonstration Shortcut through its demonstration-aware calibration.

\begin{figure}[ht!]
  \centering
  \includegraphics[width=0.7\columnwidth]{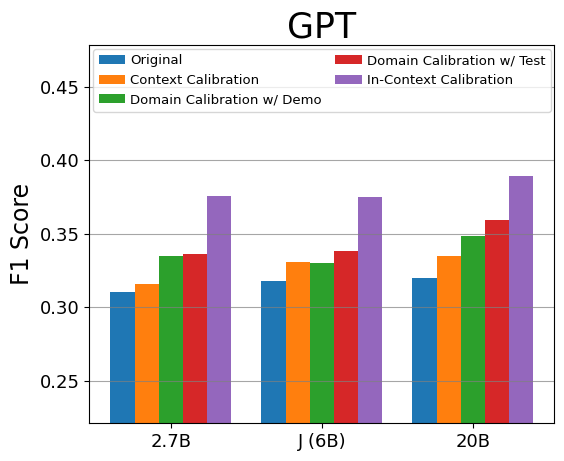} 
  \caption{Averaged Macro F1 scores for the GPT model are presented across 27 classification tasks, each featuring a permuted label space. The x-axis represents the model size. Results for the OPT and Llama2 models are provided in Appendix \ref{appendix:more_results}.}
  \label{fig:gpt_override}
\end{figure}

\subsection{Analysis with Other Input-Label Mappings}

\paragraph{Overridding Semantic Priors (Permutating Label Space)}

\citet{wei2023larger} and  \citet{kossen2023context} reveal that LLMs struggle to override semantic priors preference with input-label mappings in ICL setting. 
To test whether our proposed method helps the models override semantic priors preference by using its task learning abilities, we evaluate performance by randomly permuting the label space.
For instance, in the AGNews dataset, the original label `sports' is permuted to `world,' `business' data to `technology,' `technology' data to `sports,' and `world' data to `business' to construct a demonstration set.
The test labels are similarly permuted for evaluation.
Due to semantic priors preferences, the models tend to rely more on their pre-trained knowledge than on learning new relationships from input-label pairs, resulting in lower performance on permuted datasets.
We report the average F1 score across 27 datasets.
Figure \ref{fig:gpt_override} illustrates that the GPT model outperforms other calibration methods using In-Context Calibration.
For results related to other models, please refer to Figure \ref{fig:opt_llama2_override} in Appendix \ref{appendix:more_results}.
This indicates that In-Context Calibration's demonstration-aware calibrating is needed to diminish the model's semantic priors preferences and let it learn new tasks from demonstrations, especially those that contradict pre-trained knowledge.


\paragraph{Task Learning with different label mapping (Symbol Token)}

\begin{figure}[ht!]
  \centering
  \includegraphics[width=0.7\columnwidth]{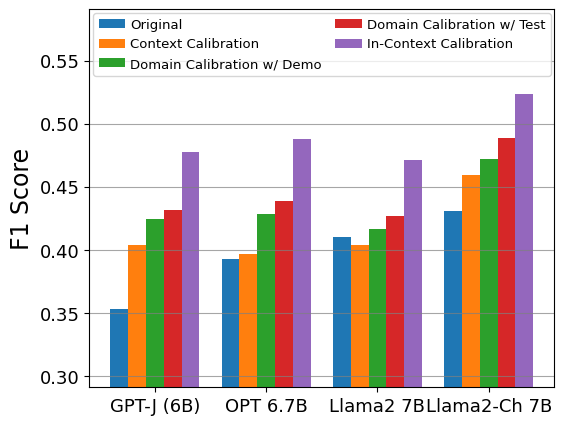} 
  \caption{Averaged Macro F1 scores for the 6-7B scale model families are presented across 27 datasets with each label space replaced by symbol tokens. The x-axis represents the model type. Results for the 13-20B scale models are available in Figure \ref{fig:symbol_large}.}
  \label{fig:symbols}
\end{figure}

\citet{pan-etal-2023-context} demonstrate that replacing the label space with symbols in the Task Learning setting leads to underperformance, attributing to their unnaturalness in pre-training stages.
To verify whether our method enhances learning ability in a more general task learning environment, we conducted experiments by mapping the label space to symbols. 
In other words, we randomly replace the label space with one of [@, \#, \$, ...] at every seed\footnote{Please refer to the Appendix \ref{appendix:impletation} for the detailed implementation.}.
We report the average F1-score across 27 datasets as experimental results. 
Figure \ref{fig:symbols} shows that our method outperforms other calibration methods, particularly demonstrating a significant improvement over the original inference.
This indicates that our method enhances LLM's task learning ability in a broader context.


\subsection{Analysis of $\lambda$ across Different Task Categories}
\label{task-wise}

\begin{table}[ht!]
\renewcommand{\arraystretch}{1.3} 
\resizebox{\columnwidth}{!}{ 
\centering
{\fontsize{9pt}{9pt}\selectfont 
\begin{tabular}{lccc}
\hline
Task & Orig. & ICC ($\lambda = 1$) & ICC ($\lambda = 0.5$) \\
\hline
Sentiment & 0.69 & 0.75 & 0.78 \\
NLI & 0.33 & 0.51 & 0.49 \\
Detection & 0.41 & 0.52 & 0.54 \\
\hline
\end{tabular}
}}
\caption{GPT-J's averaged F1 scores across different task categories with and without applying $R()$.
Orig. denotes original inference, while ICC stands for In-Context Calibration.
}
\label{table:discussion}
\end{table}

We investigated the utility of $R()$ across the different task categories, by calculating the task-wise average F1 score as shown in Table \ref{table:discussion}.
For the Sentiment and Detection task, where the semantics of each word are crucial in associating a specific label in pre-trained knowledge, applying $R()$ for calibration proves more effective.
In contrast, for NLI tasks, where the LLMs must discern the logical relationship between two input sentences, In-Context Calibration demonstrates its effectiveness over the baseline in mitigating the reliance of LLMs on semantic priors of the demonstration (as show in Figure \ref{fig:main_results}).
However, applying $R()$ in NLI tasks disrupts the order of the sentences, necessitating contextual awareness calibrating for better performance.
We reserved for additional results in Appendix \ref{appendix:full-lambda}.

\section{Conclusion}
In this paper, we introduced the term `Demonstration Shortcut', which refers to the reliance of LLMs on their pre-trained semantic priors of demonstrations for making ICL predictions.
To rectify this Demonstration Shortcut and enable the model to learn the input-label relationships from the demonstrations, we proposed a novel method, In-Context Calibration, based on the provided demonstrations.
This demonstration-aware calibration consistently yields the improved performance, regardless of model sizes or types, across various settings. 
With the introduction of In-Context Calibration, we anticipate more reliable applications of Large Language Models.

\newpage
\section*{Limitations}
In this work, we investigate the reliance of Large Language Models on their pre-trained semantic priors on demonstrations in in-context learning prediction. 
While In-Context Calibration demonstrates its effectiveness across various tasks and enhances the LLMs' task learning abilities, our experiments primarily focus on classification tasks. However, the effect of the Demonstration Shortcut might manifest differently in generation tasks. 
Further analysis and adaptation of our In-Context Calibration method for these tasks are left for future research.
Due to budget constraints, experiments with larger models (e.g., GPT4 API) and in multilingual settings were not feasible. 
Future studies with diverse settings and sufficient resources could provide a more comprehensive understanding.

Due to computational constraints, it was impractical to explore every possible $\lambda$ value for each model. 
While there may be variations across different models that remain unexplored,
we believe that the comprehensive analysis provided in the appendix will offer practical guidelines for selecting the $\lambda$ value.

\section*{Ethical Considerations}
Our work focuses on how Large Language Models utilize demonstrations in in-context learning. 
To enhance the ability of LLMs to learn input-label relationships from demonstrations, we conducted several additional inferences, requiring only minimal computational resources compared to updating model parameters. 
Additionally, we used only open-source LLMs and publicly available text classification datasets. 
Therefore, we do not concern about significant ethical issues arising from our work. 
On the contrary, we anticipate that future works could utilize our analysis to rectify harmful biases inherent in pre-trained model knowledge through demonstration-based methods.

\section*{Acknowledgements}
We appreciate Keonwoo Kim, Jaehee Kim, Yukyung Lee, and
Hyowon Cho for their invaluable comments.
We also thank POSTECH DI LAB members and anonymous reviewers for their comments on the paper.

\bibliography{custom}

\appendix
\newpage
\section{Full Description of the Articles}
\label{appendix:full_description}
\begin{figure*}[ht!]
\centering
\includegraphics[width=0.97\textwidth]{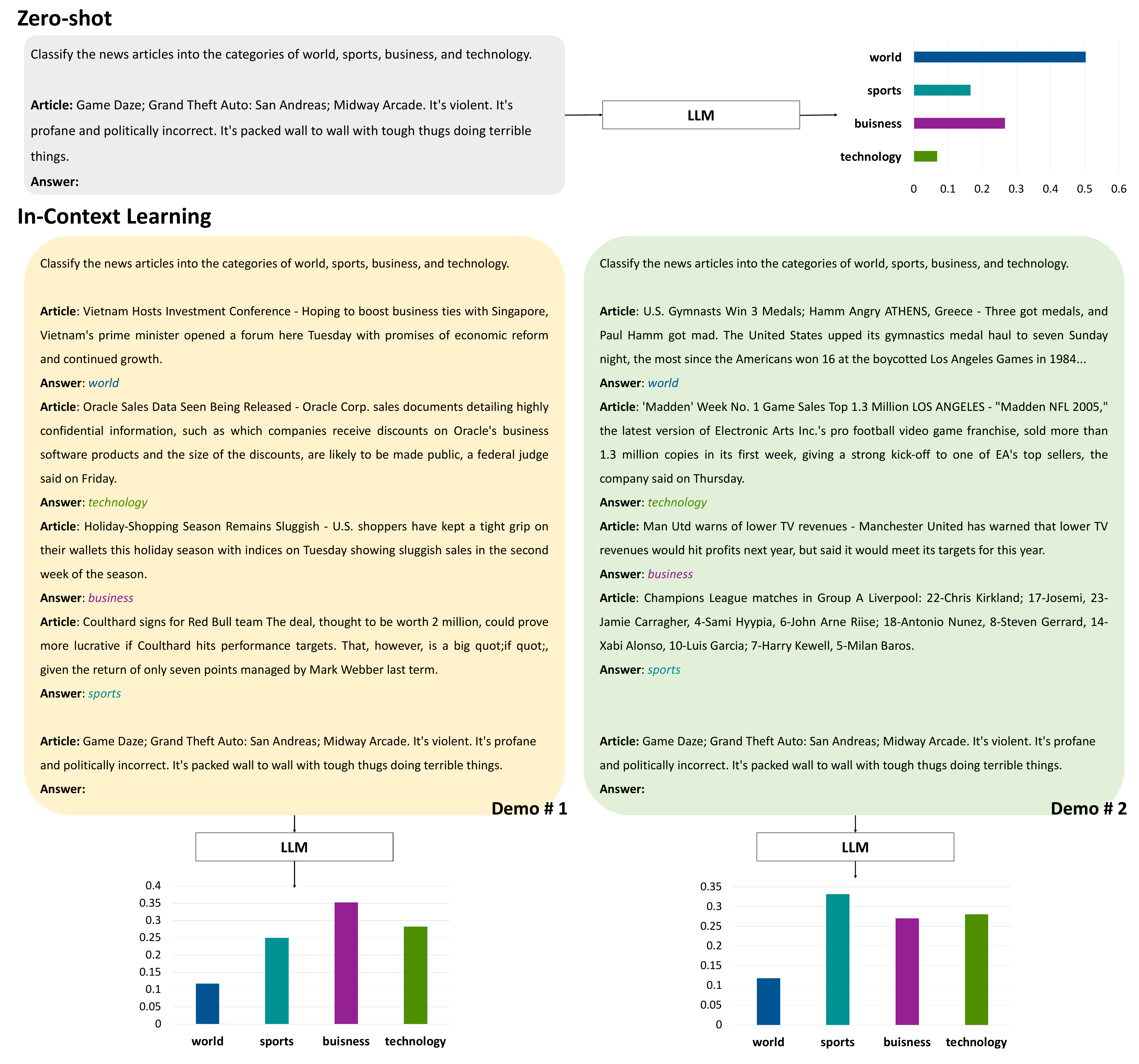}
\caption{Full description of the \textit{Demonstration Shortcut} illustrated in Figure \ref{fig:short-cut}. All articles and labels for the demonstrations and the test article were selected from the AGNews \citep{Zhang2015CharacterlevelCN} dataset. GPT-J was used in this experiment.
}
\label{fig:short-cut-full}
\end{figure*}

A full description of the articles in Figure \ref{fig:short-cut} is provided in Figure \ref{fig:short-cut-full}. 
We used the AGNews dataset \citep{Zhang2015CharacterlevelCN} for this experiment. 
In a zero-shot setting, the LLM (GPT-J) predicts the test article the world label, although the ground-truth label is technology. 
We then sampled two demonstration sets from the same training set, each having a uniform label distribution and the same label order. 
The first demonstration set predominantly features business-related semantics across all examples, while the second set leans more toward sports-related semantics. 
Despite both demonstration sets having the same uniform label distribution and identical order, the LLM fails to learn the input-label relationships, instead relying on semantic priors of demonstration to make ICL predictions.

\section{Datasets}
\label{appendix:data_info}
We use 27 text classification datasets for our experiments, most of which are widely used in existing ICL works \citep{min-etal-2022-rethinking, fei2023mitigating, pan-etal-2023-context}.
\textbf{Sentiment} task datasets include AGNews \citep{Zhang2015CharacterlevelCN}, CR \citep{hu2004mining}, financial\_phrasebank \citep{Malo2014GoodDO}, poem\_sentiment \citep{sheng-uthus-2020-investigating}, MR \citep{pangb2005exploitingclassrelationshipsforsentimentcate}, sst2 \citep{socher-etal-2013-recursive}, Subj \citep{pang2004sentimental}, and TREC \citep{hovy-etal-2001-toward};
\textbf{Natural Language Inference} task datasets include ANLI \citep{nie2019adversarial}, WNLI \citep{levesque2012winograd}, RTE \citep{dagan2005pascal}, CB \citep{de2019commitmentbank}, and SICK \citep{marelli-etal-2014-sick}
For the \textbf{Detection} task we use social\_bias\_frames \cite{sap-etal-2020-social}, tweet\_eval\_stance\_athesim, tweet\_eval\_stance\_feminist \citep{mohammad2016semeval}, tweet\_eval\_hate \citep{basile-etal-2019-semeval}, tweet\_eval\_irony \citep{van2018semeval}, tweet\_eval\_offensive \citep{zampieri2019semeval}, hate\_speech18 \citep{gibert2018hate}, ethos\_binary, ethos\_disability, ethos\_gender, ethos\_national\_origin, ethos\_race, ethos\_religion, and ethos\_violence \citep{mollas2020ethos}. 

We constructed demonstrations by sampling from the training set and using the validation set for evaluation. 
In cases where a validation set does not exist, we utilized the test set. 
For evaluation, we sampled either a maximum of 500 examples or the entire dataset size, whichever is larger.

\section{Prompt Templates}
\label{appendix:prompt}
Our natural language prompts are largely based on \citet{pmlr-v139-zhao21c} and \citet{fei2023mitigating}. We have adjusted the templates as needed to better align with each dataset's specific intent. The complete list of prompts is provided in Table \ref{table:prompt}.

\section{Implementation Details}
\label{appendix:impletation}
We set $M=20$ for the Domain Calibration implementation, following the original settings used by \citet{fei2023mitigating}.
We mainly followed the \citet{pan-etal-2023-context}'s setting for symbol token selection and incorporated additional symbol tokens to enhance generalization. 
Consequently, in each seed, every label was randomly mapped to one of the following symbols: $[``@", ``\#", ``\$", ``\%", ``\ast", ``\wedge", ``\#\#", ``\$\$", ``\%\%", ``**"]$.

\section{Random Shuffling Function}
\label{appendix:random-shuffling}

In our main experiment, we applied the random shuffling function only once to each $x_i$ to calculate $P_{R(i)}$. 
To validate whether a single shuffling process introduces randomness, we conducted additional experiments, varying the number of shuffles to 1, 5, and 10. 
We present supplementary ablation studies using GPT-J with default settings in our main experiments.

\begin{table}[h!]
\renewcommand{\arraystretch}{1.05} 
\resizebox{\columnwidth}{!}{
\centering
{\fontsize{9pt}{9pt}\selectfont 
\begin{tabular}{@{}lcc@{}}
\toprule
Method & Original ICL & TL \\
\midrule
Original Inference & 0.48 & 0.40 \\
Context Calibration & 0.47 & 0.34 \\
Domain Calibration (w/ Test) & 0.55 & 0.43 \\
ICC (1) & 0.60 & 0.49 \\
ICC (5) & 0.60 & 0.48 \\
ICC (10) & 0.60 & 0.49 \\
\bottomrule
\end{tabular}
}}
\caption{ICC stands for In-Context Calibration, and the number in (N) means shuffling number. We calculated the expectation of the randomly shuffled term when the shuffling number exceeded 1.}
\label{your-label-here}
\end{table}

The results illustrate that a single shuffle does not incur significant randomness in the estimations. 
These observations are similar to those in \citet{fei2023mitigating}, which used different random seeds to demonstrate the rapid stabilization and convergence of the sampling process.

\section{Comprehensive analysis for $\lambda$ Values}
\label{appendix:full-lambda}

\begin{table}[h!]
\renewcommand{\arraystretch}{1.3} 
\resizebox{\columnwidth}{!}{
\centering
{\fontsize{9pt}{9pt}\selectfont 
\begin{tabular}{lcc}
\hline
Method & Original & TL \\
\hline
Original Inference & 0.48 & 0.40 \\
Context Calibration & 0.47 & 0.34 \\
Domain Calibration (w/ Test)  & 0.55 & 0.43 \\
ICC ($\lambda = 0 $) & 0.56 & 0.46 \\
ICC ($\lambda = 0.25 $) & 0.56 & 0.45 \\
ICC ($\lambda = 0.5 $)& 0.60 & 0.49 \\
ICC ($\lambda = 0.75 $) & 0.57 & 0.47 \\
ICC ($\lambda = 1 $) & 0.58 & 0.50 \\
\hline
\end{tabular}
}}
\caption{Average Macro F1 scores across
27 datasets for both Original ICL Task (Original) and Task Learning (TL) with different $\lambda$ values using GPT-J. ICC stands for In-Context Calibration.}
\label{table:full-lambda-GPT-J}
\end{table}

We present the complete results for various $\lambda$ values of GPT-J in Table \ref{table:full-lambda-GPT-J}.
GPT-J performs best with an Original ICL Task at 0.5 and the higher $\lambda$ values lead to higher task learning ability. 

\begin{table}[h!]
\renewcommand{\arraystretch}{1.3} 
\resizebox{\columnwidth}{!}{
\centering
{\fontsize{14pt}{14pt}\selectfont 
\begin{tabular}{@{}lcccc@{}}
\toprule
 & \multicolumn{2}{c}{Original} & \multicolumn{2}{c}{TL} \\
\cmidrule(r){2-3} \cmidrule(r){4-5}
 & OPT-6.7B & Llama2-7B & OPT-6.7B & Llama2-7B \\
\midrule
Original Inference & 0.48 & 0.48 & 0.38 & 0.45 \\
Context Calibration & 0.48 & 0.48 & 0.35 & 0.43 \\
Domain Calibration (w/ Test) & 0.52 & 0.55 & 0.43 & 0.49 \\
ICC ($\lambda = 0$) & 0.55 & 0.56 & 0.46 & 0.51 \\
ICC ($\lambda = 0.25$) & 0.55 & 0.58 & 0.47 & 0.51 \\
ICC ($\lambda = 0.5$) & 0.57 & 0.61 & 0.49 & 0.52 \\
ICC ($\lambda = 0.75$) & 0.56 & 0.60 & 0.47 & 0.51 \\
ICC ($\lambda = 1.0$) & 0.57 & 0.59 & 0.49 & 0.53 \\
\bottomrule
\end{tabular}
}
}
\caption{Average Macro F1 scores across
27 datasets for both Original ICL Task (Original) and Task Learning (TL)  with different $\lambda$ values for OPT-6.7B and Llama2-7B. ICC stands for In Context Calibration.}
\label{table:full-lambda-GPT-J-task-wise}
\end{table}

To further analyze the comprehensive impact of $\lambda$, we conducted supplementary studies using OPT-6.7B and Llama2-7B, which have similar sizes to GPT-J, exploring broader $\lambda$ values (Table \ref{table:full-lambda-GPT-J-task-wise}).
As with GPT-J’s $\lambda$ value, OPT-6.7B and Llama2-7B also perform best with an Original ICL Task at 0.5 (OPT-6.7B also performs best at the value of 1.0).
In line with our primary findings (Table \ref{table:discussion}), a higher $\lambda$ value enhances task learning capabilities, leading to less grammatical information corruption in the original input sentences.
We also analyze the task-wise value as in Table \ref{table:task-wise-others}. 
Consistent across all models, a higher $\lambda$ improves performance in the NLI task, corroborating our initial findings. 
Similar to the results seen in \citet{fei2023mitigating}, the semantics of words play a critical role in Sentiment and Detection tasks (where a lower $\lambda$ still shows comparative performance, despite the lost of grammatical information), as the language model's dependence on the label's semantics is significant in these tasks.
Therefore, (1) we recommend starting with a $\lambda$ value of 0.5 in In-Context Calibration and adjusting it based on the task-wise experimental results obtained from the validation set in the Original ICL Task and (2) search for a value with 0.5 or higher in the Task Learning setting. 
However, due to computational constraints, performing a grid search across all models for every $\lambda$ value was impractical, so we opted for $\lambda$ of 0.5 in the main experiment for all models for efficiency.

\section{Detailed Results}
\label{appendix:more_results}

We demonstrate the results of the OPT, GPT, Llama2, and Llama2-Chat models in both the Original ICL Task and the Task Learning setting across Table \ref{table:OPT_original_ICL} to \ref{table:llama2-chat_TL}. 
For models above 50B scale, please refer to Table \ref{table:LLM_original_ICL} and \ref{table:LLM_TL}. 
We average the performance across a total of 5 seeds, recording mean values and standard deviations. `Orig.' denotes original inference, `CC' refers to Context Calibration, `DC' stands for Domain Calibration, and `ICC' represents In-Context Calibration.
Furthermore, our results indicate that In-Context Calibration outperforms the baselines in most tasks.

\begin{figure}[h!]
\centering
\includegraphics[width=0.4\textwidth]{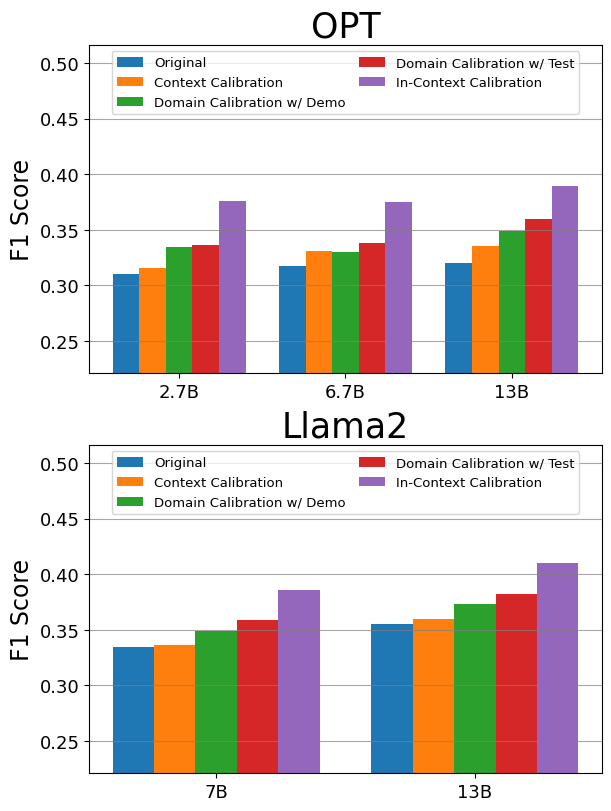} 
\caption{Averaged Macro F1 scores for the OPT (Top Graph) and Llama2 (Bottom graph) model are presented across 27 classification tasks, each featuring a permuted label space.
The x-axis represents the model size.}
\label{fig:opt_llama2_override}
\end{figure}
Figure \ref{fig:opt_llama2_override} presents experimental results for OPT and Llama2 under the same setting as figure \ref{fig:gpt_override}. 
We demonstrate that our method is effective across various model types and sizes. 
Specifically, OPT shows a 19\% performance increase compared to the original inference, while Llama2 exhibits a 15\% improvement.

\begin{figure}[h!]
\centering
\includegraphics[width=0.4\textwidth]{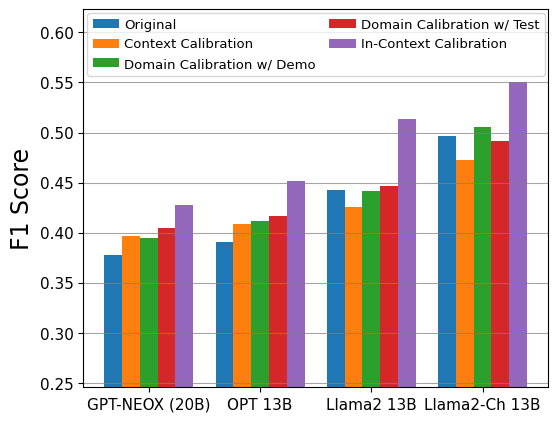} 
\caption{Averaged Macro F1 scores for the 13-20B scale model families are presented across 27 datasets with each label space replaced by symbol tokens. The x-axis represents the model type.}
\label{fig:symbol_large}
\end{figure}

In Figure \ref{fig:symbol_large}, we present the performance of LLMs when labels are mapped to symbols. 
In-Context Calibration significantly improves performance across different model types and sizes, highlighting the consistent enhancement of task learning ability.

\section{Adding More In-Context Examples (K=8/12/16)}
\label{appendix:more_k}

\begin{figure}[h!]
  \centering
  \includegraphics[width=0.4\textwidth]{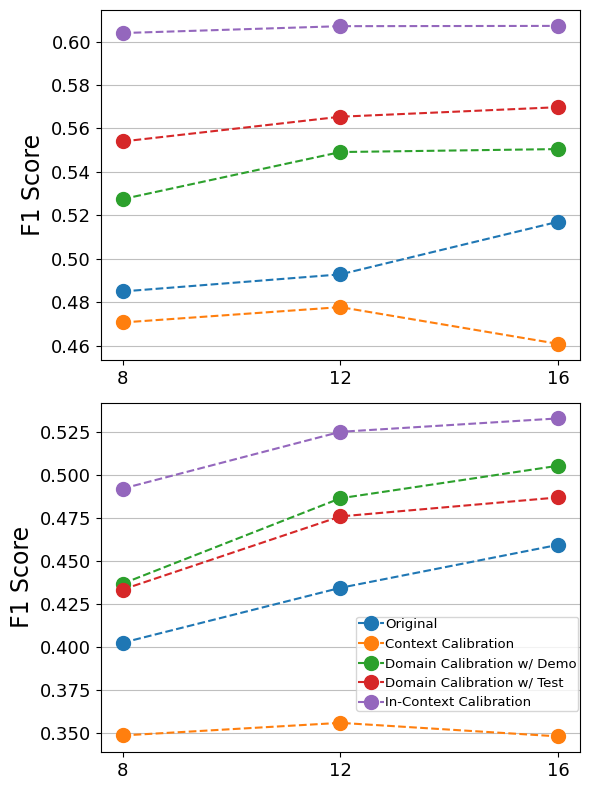} 
  \caption{The top graph depicts the average Macro F1 score for the Original ICL Task across 27 datasets for GPT-J. The bottom graph plots the average Macro F1 score for the Task Learning setting. In both graphs, the x-axis represents the number of demonstrations.}
  \label{fig:more-k}
\end{figure}
We study the effect of adding more in-context examples by evaluating the GPT-J. 
As shown in Figure \ref{fig:more-k}, In-Context Calibration achieves approximately a 6\% higher performance than other calibration methods in the Original ICL Task. 
Similar to \citet{pan-etal-2023-context}, Task Learning performance further improves as the number of demonstrations increases.
In-Context Calibration consistently outperforms the baselines, whereas Context Calibration lags behind the original inference.
In conclusion, In-Context Calibration consistently demonstrates superior performance in both the Original ICL Task and Task Learning settings, irrespective of the number of demonstrations.


\newpage

\renewcommand{\arraystretch}{1.2}

\begin{table*}[h!]
\small
\centering
\begin{tabular}{lll}
\hline
\textbf{Dataset}           & \textbf{Natural Language Template}                                       & \textbf{Label Space}      \\ \hline
\textbf{Sentiment Task}           &                                        &                 \\ \hline
sst2, MR, CR       & Review: [INPUT]                                         & positive, negative                  \\ 
                           & Sentiment: [LABEL]                                      &                                     \\ \hline
financial\_phrasebank       & Sentence: [INPUT]                                       & positive, negative,         \\ 
                           & Sentiment: [LABEL]                                      & neutral                \\ \hline
poem\_sentiment             & Verse text: [INPUT]                                     & positive, negative,  \\
                           & Sentiment: [LABEL]                                      & neutral                \\
\hline
\end{tabular}
\caption{Prompt templates used in our experiments are primarily adapted from the work of \citet{pmlr-v139-zhao21c} and \citet{fei2023mitigating}. In the Task Learning setting, each label is randomly mapped to a string number corresponding to the size of the Label Space (i.e., from 0 to len(Label Space) - 1) for each seed.}
\label{table:prompt}
\end{table*}

\renewcommand{\arraystretch}{0.85}

\begin{table*}[ht!]
\centering
\begin{tabular}{@{}lcccccc@{}}
\toprule
 & \multicolumn{2}{c}{Sentiment} & \multicolumn{2}{c}{NLI} & \multicolumn{2}{c}{Detection} \\
\cmidrule(r){2-3} \cmidrule(r){4-5} \cmidrule(r){6-7}
 & OPT-6.7B & Llama2-7B & OPT-6.7B & Llama2-7B & OPT-6.7B & Llama2-7B \\
\midrule
Original Inference & 0.674 & 0.720 & 0.342 & 0.478 & 0.457 & 0.371 \\
Contextual Calibration & 0.685 & 0.735 & 0.326 & 0.515 & 0.465 & 0.423 \\
Domain Calibration  w/ Test & 0.734 & 0.752 & 0.316 & 0.496 & 0.524 & 0.523 \\
ICC ($\lambda = 0$) & 0.764 & 0.774 & 0.338 & 0.454 & 0.543 & 0.523 \\
ICC ($\lambda = 0.25$) & 0.764 & 0.796 & 0.337 & 0.511 & 0.522 & 0.555 \\
ICC ($\lambda = 0.5$) & 0.749 & 0.792 & 0.398 & 0.551 & 0.552 & 0.558 \\
ICC ($\lambda = 0.75$) & 0.751 & 0.782 & 0.428 & 0.557 & 0.532 & 0.564 \\
ICC ($\lambda = 1.0$) & 0.749 & 0.775 & 0.445 & 0.547 & 0.530 & 0.532 \\
\bottomrule
\end{tabular}
\caption{OPT-6.7B and Llama2-7B's averaged F1 scores across different task categories using various $\lambda$ value. ICC stands for In Context Calibration.}
\label{table:task-wise-others}
\end{table*}

\renewcommand{\arraystretch}{1.0}

\begin{table*}[h!]
\centering
\resizebox{0.6\textwidth}{!}{
\begin{tabular}{cccccccc}
\hline
Dataset & Size & \multicolumn{5}{c}{Method} \\ \hline
 Sentiment Task &  &  Orig.  &  CC  & DC w/Demo & DC w/Test  &  ICC  \\ \hline
\multirow{3}{*}{AGNews} & 2.7B & 0.298 (0.072) &   0.669 (0.117)   &   0.676 (0.676)   &  0.803 (0.018)    &  0.811 (0.024)   \\ \cline{2-7}
& 6.7B & 0.582 (0.104) &  0.716 (0.108)   &   0.677 (0.677)   &  0.765 (0.061)    &  0.811 (0.023)    \\ \cline{2-7}
& 13B & 0.722 (0.070) &   0.717 (0.104)   &   0.735 (0.735)   &  0.796 (0.045)    &  0.828 (0.051)   \\ \hline
\multirow{3}{*}{CR} & 2.7B & 0.888 (0.031) &   0.892 (0.021)   &   0.911 (0.911)   &  0.915 (0.008)    &  0.908 (0.018)   \\ \cline{2-7}
& 6.7B & 0.860 (0.042) &  0.883 (0.024)   &   0.911 (0.911)   &  0.921 (0.007)    &  0.921 (0.005)    \\ \cline{2-7}
& 13B & 0.860 (0.082) &   0.912 (0.013)   &   0.921 (0.921)   &  0.920 (0.013)    &  0.926 (0.006)   \\
\hline
\end{tabular}}
\caption{\label{table:OPT_original_ICL}Macro F1-score across 27 datasets for OPT model family under the Original ICL Task.}
\end{table*}

\begin{table*}[h!]
\centering
\resizebox{0.6\textwidth}{!}{
\begin{tabular}{cccccccc}
\hline
Dataset & Size & \multicolumn{5}{c}{Method} \\ \hline
 Sentiment Task &  &  Orig.  &  CC  & DC w/Demo & DC w/Test  &  ICC  \\ \hline
\multirow{3}{*}{AGNews} & 2.7B & 0.635 (0.092) &   0.724 (0.041)   &   0.596 (0.596)   &  0.771 (0.043)    &  0.727 (0.039)   \\ \cline{2-7}
& J (6B) & 0.755 (0.129) &  0.787 (0.073)   &   0.726 (0.726)   &  0.755 (0.046)    &  0.867 (0.012)    \\ \cline{2-7}
& 20B & 0.640 (0.066) &   0.826 (0.031)   &   0.798 (0.798)   &  0.813 (0.014)    &  0.818 (0.030)   \\ \hline
\multirow{3}{*}{CR} & 2.7B & 0.744 (0.106) &   0.797 (0.095)   &   0.857 (0.857)   &  0.878 (0.032)    &  0.887 (0.019)   \\ \cline{2-7}
& J (6B) & 0.869 (0.052) &  0.885 (0.026)   &   0.904 (0.904)   &  0.911 (0.002)    &  0.912 (0.004)    \\ \cline{2-7}
& 20B & 0.895 (0.031) &   0.919 (0.010)   &   0.914 (0.914)   &  0.918 (0.008)    &  0.919 (0.005)   \\
\hline
\end{tabular}}
\caption{ Macro F1-score across 27 datasets for GPT model family under the Original ICL Task.}
\end{table*}

\begin{table*}[h!]
\centering
\resizebox{0.6\textwidth}{!}{
\begin{tabular}{cccccccc}
\hline
Dataset & Size & \multicolumn{5}{c}{Method} \\ \hline
 Sentiment Task &  &  Orig.  &  CC  & DC w/Demo & DC w/Test  &  ICC  \\ \hline
\multirow{2}{*}{AGNews} & 7B & 0.803 (0.010) &   0.770 (0.013)   &   0.753 (0.753)   &  0.759 (0.016)    &  0.822 (0.018)   \\ \cline{2-7}
& 13B & 0.741 (0.046) &   0.733 (0.035)   &   0.730 (0.730)   &  0.730 (0.026)    &  0.764 (0.047)   \\ \hline
\multirow{2}{*}{CR} & 7B & 0.892 (0.039) &   0.926 (0.010)   &   0.929 (0.929)   &  0.927 (0.007)    &  0.938 (0.013)   \\ \cline{2-7}
& 13B & 0.928 (0.023) &   0.934 (0.013)   &   0.932 (0.932)   &  0.931 (0.015)    &  0.936 (0.015)   \\ \hline
\multirow{2}{*}{financial\_phrasebank} & 7B & 0.665 (0.192) &   0.426 (0.066)   &   0.515 (0.515)   &  0.588 (0.102)    &  0.720 (0.082)   \\ \cline{2-7}
& 13B & 0.631 (0.141) &   0.598 (0.066)   &   0.744 (0.744)   &  0.752 (0.037)    &  0.772 (0.029)   \\
\hline
\end{tabular}}
\caption{ Macro F1-score across 27 datasets for Llama2 model family under the Original ICL Task.}
\end{table*}

\begin{table*}[h!]
\centering
\resizebox{0.6\textwidth}{!}{
\begin{tabular}{cccccccc}
\hline
Dataset & Size & \multicolumn{5}{c}{Method} \\ \hline
 Sentiment Task &  &  Orig.  &  CC  & DC w/Demo & DC w/Test  &  ICC  \\ \hline
\multirow{2}{*}{AGNews} & 7B & 0.734 (0.059) &   0.774 (0.041)   &   0.773 (0.773)   &  0.767 (0.040)    &  0.784 (0.042)   \\ \cline{2-7}
& 13B & 0.765 (0.018) &   0.745 (0.019)   &   0.778 (0.778)   &  0.772 (0.029)    &  0.774 (0.030)   \\ \hline
\multirow{2}{*}{CR} & 7B & 0.926 (0.009) &   0.920 (0.007)   &   0.924 (0.924)   &  0.927 (0.011)    &  0.930 (0.008)   \\ \cline{2-7}
& 13B & 0.919 (0.021) &   0.928 (0.012)   &   0.928 (0.928)   &  0.928 (0.012)    &  0.937 (0.010)   \\ \hline
\multirow{2}{*}{financial\_phrasebank} & 7B & 0.869 (0.052) &   0.574 (0.058)   &   0.759 (0.759)   &  0.774 (0.049)    &  0.877 (0.040)   \\ \cline{2-7}
& 13B & 0.720 (0.119) &   0.708 (0.075)   &   0.806 (0.806)   &  0.790 (0.047)    &  0.829 (0.041)   \\
\hline
\end{tabular}}
\caption{ Macro F1-score across 27 datasets for Llama2-chat model family under the Original ICL Task.}
\end{table*}

\begin{table*}[h!]
\centering
\resizebox{0.6\textwidth}{!}{
\begin{tabular}{cccccccc}
\hline
Dataset & Size & \multicolumn{5}{c}{Method} \\ \hline
 Sentiment Task &  &  Orig.  &  CC  & DC w/Demo & DC w/Test  &  ICC  \\ \hline
\multirow{3}{*}{AGNews} & 2.7B & 0.193 (0.058) &   0.167 (0.037)   &   0.187 (0.187)   &  0.195 (0.043)    &  0.261 (0.032)   \\ \cline{2-7}
& 6.7B & 0.184 (0.068) &  0.143 (0.043)   &   0.228 (0.228)   &  0.226 (0.045)    &  0.305 (0.030)    \\ \cline{2-7}
& 13B & 0.205 (0.071) &   0.155 (0.048)   &   0.193 (0.193)   &  0.294 (0.029)    &  0.311 (0.047)   \\ \hline
\multirow{3}{*}{CR} & 2.7B & 0.838 (0.055) &   0.738 (0.090)   &   0.865 (0.865)   &  0.902 (0.016)    &  0.905 (0.009)   \\ \cline{2-7}
& 6.7B & 0.587 (0.255) &  0.507 (0.080)   &   0.844 (0.844)   &  0.867 (0.026)    &  0.847 (0.069)    \\ \cline{2-7}
& 13B & 0.687 (0.180) &   0.841 (0.097)   &   0.875 (0.875)   &  0.900 (0.019)    &  0.884 (0.037)   \\
\hline
\end{tabular}}
\caption{ Macro F1-score across 27 datasets for OPT model family under the Task Learning setting.}
\end{table*}

\begin{table*}[h!]
\centering
\resizebox{0.6\textwidth}{!}{
\begin{tabular}{cccccccc}
\hline
Dataset & Size & \multicolumn{5}{c}{Method} \\ \hline
 Sentiment Task &  &  Orig.  &  CC  & DC w/Demo & DC w/Test  &  ICC  \\ \hline
\multirow{3}{*}{AGNews} & 2.7B & 0.215 (0.044) &   0.151 (0.039)   &   0.299 (0.299)   &  0.349 (0.063)    &  0.396 (0.055)   \\ \cline{2-7}
& J (6B) & 0.205 (0.032) &  0.197 (0.048)   &   0.238 (0.238)   &  0.267 (0.046)    &  0.314 (0.036)    \\ \cline{2-7}
& 20B & 0.229 (0.030) &   0.206 (0.041)   &   0.239 (0.239)   &  0.281 (0.048)    &  0.314 (0.034)   \\ \hline
\multirow{3}{*}{CR} & 2.7B & 0.603 (0.176) &   0.770 (0.077)   &   0.772 (0.772)   &  0.849 (0.050)    &  0.835 (0.043)   \\ \cline{2-7}
& J (6B) & 0.695 (0.138) &  0.618 (0.125)   &   0.879 (0.879)   &  0.868 (0.057)    &  0.919 (0.003)    \\ \cline{2-7}
& 20B & 0.727 (0.159) &   0.869 (0.044)   &   0.860 (0.860)   &  0.886 (0.022)    &  0.889 (0.025)   \\
\hline
\end{tabular}}
\caption{Macro F1-score across 27 datasets for GPT model family under the Task Learning setting.}
\end{table*}

\begin{table*}[h!]
\centering
\resizebox{0.6\textwidth}{!}{
\begin{tabular}{cccccccc}
\hline
Dataset & Size & \multicolumn{5}{c}{Method} \\ \hline
 Sentiment Task &  &  Orig.  &  CC  & DC w/Demo & DC w/Test  &  ICC  \\ \hline
\multirow{2}{*}{AGNews} & 7B & 0.377 (0.045) &   0.317 (0.068)   &   0.344 (0.344)   &  0.334 (0.088)    &  0.462 (0.019)   \\ \cline{2-7}
& 13B & 0.381 (0.053) &   0.315 (0.054)   &   0.320 (0.320)   &  0.297 (0.064)    &  0.488 (0.026)   \\ \hline
\multirow{2}{*}{CR} & 7B & 0.920 (0.016) &   0.802 (0.116)   &   0.906 (0.906)   &  0.928 (0.010)    &  0.932 (0.005)   \\ \cline{2-7}
& 13B & 0.623 (0.163) &   0.569 (0.204)   &   0.774 (0.774)   &  0.821 (0.041)    &  0.823 (0.061)   \\ \hline
\multirow{2}{*}{financial\_phrasebank} & 7B & 0.346 (0.109) &   0.408 (0.083)   &   0.398 (0.398)   &  0.383 (0.062)    &  0.418 (0.093)   \\ \cline{2-7}
& 13B & 0.367 (0.095) &   0.384 (0.071)   &   0.395 (0.395)   &  0.368 (0.073)    &  0.434 (0.101)   \\
\hline
\end{tabular}}
\caption{ Macro F1-score across 27 datasets for Llama2 model family under the Task Learning setting.}
\end{table*}

\begin{table*}[h!]
\centering
\resizebox{0.6\textwidth}{!}{
\begin{tabular}{cccccccc}
\hline
Dataset & Size & \multicolumn{5}{c}{Method} \\ \hline
 Sentiment Task &  &  Orig.  &  CC  & DC w/Demo & DC w/Test  &  ICC  \\ \hline
\multirow{2}{*}{AGNews} & 7B & 0.315 (0.108) &   0.330 (0.044)   &   0.263 (0.263)   &  0.320 (0.062)    &  0.391 (0.042)   \\ \cline{2-7}
& 13B & 0.496 (0.134) &   0.413 (0.095)   &   0.479 (0.479)   &  0.374 (0.087)    &  0.528 (0.138)   \\ \hline
\multirow{2}{*}{CR} & 7B & 0.690 (0.160) &   0.525 (0.190)   &   0.609 (0.609)   &  0.690 (0.144)    &  0.723 (0.114)   \\ \cline{2-7}
& 13B & 0.841 (0.080) &   0.713 (0.202)   &   0.848 (0.848)   &  0.840 (0.093)    &  0.853 (0.083)   \\ \hline
\multirow{2}{*}{financial\_phrasebank} & 7B & 0.218 (0.072) &   0.241 (0.112)   &   0.344 (0.344)   &  0.380 (0.103)    &  0.388 (0.064)   \\ \cline{2-7}
& 13B & 0.467 (0.110) &   0.462 (0.080)   &   0.474 (0.474)   &  0.531 (0.085)    &  0.488 (0.068)   \\
\hline
\end{tabular}}
\caption{\label{table:llama2-chat_TL} Macro F1-score across 27 datasets for Llama2-chat model family under the Task Learning setting.}
\end{table*}

\begin{table*}[h!]
\centering
\resizebox{0.6\textwidth}{!}{
\begin{tabular}{cccccccc}
\hline
Dataset & Size & \multicolumn{5}{c}{Method} \\ \hline
 Sentiment Task &  &  Orig.  &  CC  & DC w/Demo & DC w/Test  &  ICC  \\ \hline
\multirow{3}{*}{AGNews} & OPT 66B & 0.793 (0.058) &   0.820 (0.049)   &   0.761 (0.761)   &  0.794 (0.098)    &  0.827 (0.073)   \\ \cline{2-7}
& Llama2 70B & 0.885 (0.016) &  0.893 (0.014)   &   0.897 (0.897)   &  0.900 (0.008)    &  0.891 (0.014)    \\ \cline{2-7}
& Llama2-ch 70B & 0.806 (0.035) &   0.782 (0.051)   &   0.804 (0.804)   &  0.819 (0.042)    &  0.813 (0.034)   \\ \hline
\multirow{3}{*}{CR} & OPT 66B & 0.852 (0.045) &   0.916 (0.012)   &   0.917 (0.917)   &  0.922 (0.006)    &  0.922 (0.009)   \\ \cline{2-7}
& Llama2 70B & 0.918 (0.039) &  0.926 (0.025)   &   0.928 (0.928)   &  0.933 (0.011)    &  0.933 (0.007)    \\ \cline{2-7}
& Llama2-ch 70B & 0.927 (0.003) &   0.927 (0.004)   &   0.927 (0.927)   &  0.927 (0.004)    &  0.928 (0.003)   \\
\hline
\end{tabular}}
\caption{\label{table:LLM_original_ICL} Macro F1-score across 27 datasets for over 50B scale LLMs under the Original ICL Task setting.}
\end{table*}

\begin{table*}[h!]
\centering
\resizebox{0.6\textwidth}{!}{
\begin{tabular}{cccccccc}
\hline
Dataset & Size & \multicolumn{5}{c}{Method} \\ \hline
 Sentiment Task &  &  Orig.  &  CC  & DC w/Demo & DC w/Test  &  ICC  \\ \hline
\multirow{3}{*}{AGNews} & OPT 66B & 0.172 (0.068) &   0.198 (0.071)   &   0.157 (0.157)   &  0.159 (0.033)    &  0.336 (0.027)   \\ \cline{2-7}
& Llama2 70B & 0.745 (0.113) &  0.744 (0.107)   &   0.751 (0.751)   &  0.739 (0.107)    &  0.785 (0.065)    \\ \cline{2-7}
& Llama2-ch 70B & 0.666 (0.098) &   0.655 (0.080)   &   0.683 (0.683)   &  0.667 (0.089)    &  0.705 (0.047)   \\ \hline
\multirow{3}{*}{CR} & OPT 66B & 0.768 (0.156) &   0.853 (0.047)   &   0.891 (0.891)   &  0.897 (0.037)    &  0.913 (0.019)   \\ \cline{2-7}
& Llama2 70B & 0.720 (0.261) &  0.722 (0.238)   &   0.803 (0.803)   &  0.811 (0.151)    &  0.805 (0.147)    \\ \cline{2-7}
& Llama2-ch 70B & 0.912 (0.019) &   0.864 (0.044)   &   0.917 (0.917)   &  0.916 (0.014)    &  0.913 (0.016)   \\
\hline
\end{tabular}}
\caption{\label{table:LLM_TL} Macro F1-score across 27 datasets for over 50B scale LLMs under the Task Learning setting.}
\end{table*}

\end{document}